\documentclass[10pt,twocolumn,letterpaper]{article}

\usepackage{iccv}
\usepackage{times}
\usepackage{epsfig}
\usepackage{graphicx}
\usepackage{amsmath}
\usepackage{amssymb}
\usepackage{tabu}
\usepackage{multirow}

\usepackage[breaklinks=true,bookmarks=false]{hyperref}

\iccvfinalcopy 

\def\iccvPaperID{} 

\begin{document}

\title{Diff-Retinex: Rethinking Low-light Image Enhancement with A Generative Diffusion Model}

\author{Xunpeng Yi$^1$, Han Xu$^1$, Hao Zhang, Linfeng Tang, and Jiayi Ma\thanks{Corresponding author}\\
Electronic Information School, Wuhan University, Wuhan 430072, China\qquad\\
{\tt\small \{yixunpeng, xu\_han\}@whu.edu.cn,\,\{zhpersonalbox, linfeng0419, jyma2010\}@gmail.com}
}

\maketitle
\ificcvfinal\thispagestyle{empty}\fi
\footnotetext[1]{These authors contributed equally to this work.}

\begin{abstract}
   In this paper, we rethink the low-light image enhancement task and propose a physically explainable and generative diffusion model for low-light image enhancement, termed as \textit{Diff-Retinex}. We aim to integrate the advantages of the physical model and the generative network. Furthermore, we hope to supplement and even deduce the information missing in the low-light image through the generative network. Therefore, Diff-Retinex formulates the low-light image enhancement problem into Retinex decomposition and conditional image generation. In the Retinex decomposition, we integrate the superiority of attention in Transformer and meticulously design a Retinex Transformer decomposition network (TDN) to decompose the image into illumination and reflectance maps. Then, we design multi-path generative diffusion networks to reconstruct the normal-light Retinex probability distribution and solve the various degradations in these components respectively, including dark illumination, noise, color deviation, loss of scene contents, \textit{etc}. Owing to generative diffusion model, Diff-Retinex puts the restoration of low-light subtle detail into practice. Extensive experiments conducted on real-world low-light datasets qualitatively and quantitatively demonstrate the effectiveness, superiority, and generalization of the proposed method.
\end{abstract}

\section{Introduction}
\begin{figure}[!t]
\centering
\includegraphics[width=0.97\linewidth]{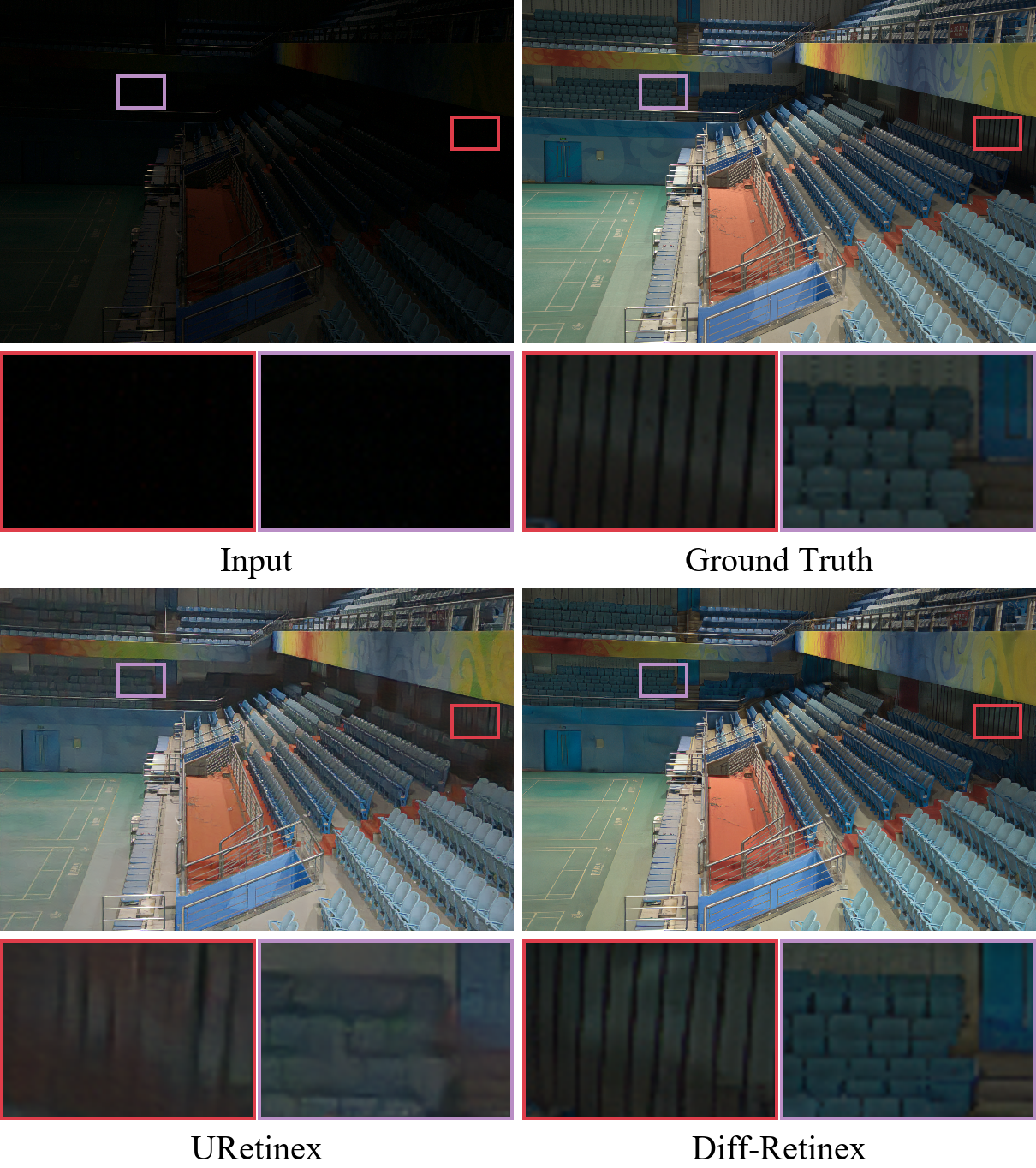}
\caption{Example of URetinex~\cite{wu2022uretinex} and Diff-Retinex for LLIE. Diff-Retinex can repair some missing scene contents by rethinking LLIE through the generative diffusion model.}
\label{fig:1}
\end{figure}

Images taken in low-light scenes are usually affected by a variety of degradations, such as indefinite noise, low contrast, variable color deviation, \textit{etc}. Among these degradations, the loss of scene structures is the most thorny. As shown in Fig.~\ref{fig:1}, the loss of scene structures is not only limited to affecting the visual effect but also reducing the amount of information. Image enhancement is an effective approach to reduce the interference of degradations on human perception and subsequent vision tasks, and finally present high-quality images.

To handle these degradations, many low-light image enhancement (LLIE) methods have been proposed~\cite{ liu2021underexposed, ma2022practical}. Besides, a series of research on contrast enhancement, noise removal, and texture preservation have been carried out. The mainstream LLIE methods can be roughly divided into traditional~\cite{ueng1995gamma, fu2016fusion} and learning-based methods~\cite{lore2017llnet, li2018lightennet, ma2022toward, ma2023bilevel, liu2022learning}. Traditional algorithms are usually based on the image prior or simple physical models. For instance, gray transform~\cite{ueng1995gamma, lee2013contrast} and histogram equalization~\cite{han2010novel, singh2014image} adjust the intensity distribution by linear or nonlinear means. The Retinex model~\cite{kong2021low, li2018structure, ma2022low} decomposes the image into illumination and reflectance images, and the problem is solved with traditional optimization methods. However, these methods are also subject to manual design and optimization-driven efficiency. They usually suffer from poor generalization and robustness, limiting the application scope of these methods.

To address these drawbacks, deep learning is utilized to construct the complex mapping from low-light to normal-light image~\cite{lore2017llnet, wang2023ultra}. Some methods entirely regard low-light image enhancement as a restoration task through an overall fitting, lacking theoretical support and interpretability of physical models. Compared with physical model-based methods, they usually exhibit less targeted enhancement performance, manifested as uneven illumination, non-robustness to noise, \textit{etc}. The main cause is the shortage of specific definitions of some degradations and targeted processing for them. The physical model-based methods decompose the image into components with physical significance. Then, specific processing is conducted on the components for more targeted enhancement.

However, existing methods hardly escape the essence of fitting. More concretely, the distorted scene can be better rendered through denoising in the existing methods while the missing scene content cannot be repaired. Take Fig.~\ref{fig:1} as an example, the state-of-the-art method (URetinex~\cite{wu2022uretinex}) cannot restore the weak and missing details and even aggravates the information distortion to a certain extent. To solve this drawback and considering that LLIE is a process of recovering a normal-light image with the guidance of the low-light image, we rethink LLIE with a generative diffusion model. We aim to recover or even reason out the weak, even the lost information in the original low-light image. Thus, LLIE is regarded as not only a restoration fitting function but also an image generation task with the condition. As for the generative model, generative adversarial networks (GAN)~\cite{wang2017generative, zhu2018generative} train a generator and a discriminator in an adversarial mechanism. However, they suffer from training instability, resulting in problems such as mode collapse, non-convergence, and exploding or vanishing gradients. Moreover, the GAN-based LLIE methods also have the problem of directly generating the normal-light image through an overall fitting, lacking physical interpretability as mentioned before.

To this end, we propose a physically explainable and generative model for low-light image enhancement, termed as \textit{Diff-Retinex}. We aim to integrate the advantages of the physical model and the generative network. Thus, Diff-Retinex formulates the low-light image enhancement problem to Retinex decomposition and conditional image generation. In the Retinex decomposition, we integrate the characteristics of Transformer~\cite{liu2021swin, zamir2022restormer} and meticulously design a Retinex Transformer decomposition network (TDN) to improve the decomposition applicability. TDN decomposes the image into illumination and reflectance maps. Then, we design generative diffusion-based networks to solve the various degradations in these components respectively, including dark illumination, noise, color deviation, loss of scene contents, \textit{etc}. The main contributions are summarized as:
\begin{itemize}
  \item[-] We rethink low-light image enhancement from the perspective of conditional image generation. Rather than being limited to enhancing the original low-quality information, we propose a generative Retinex framework to further compensate for content loss and color deviation caused by low light.
  \item[-] Considering the issues of decomposition in Retinex models, we propose a novel Transformer decomposition network. It can take full advantage of the attention and layer dependence to efficiently decompose images, even for images of high resolutions.
  \item[-] To the best of our knowledge, it is the first study that applies the diffusion model with Retinex model for low-light image enhancement. The diffusion model is applied to guide the multi-path adjustments of illumination and reflectance maps for better performance.
\end{itemize}

\begin{figure*}[t]
\centering
\includegraphics[width=0.98\linewidth]{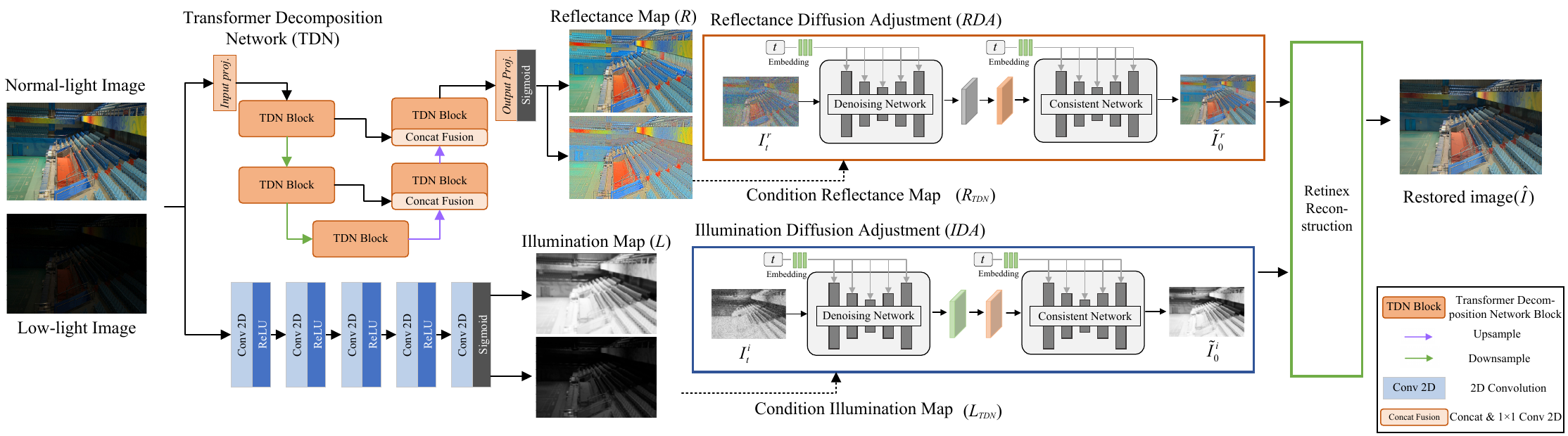}
   \caption{Overall framework of Diff-Retinex. It contains three detachable modules, \textit{i.e.}, Transformer Decomposition Network (TDN), Reflectance Diffusion Adjustment (RDA), and Illumination Diffusion Adjustment (IDA).} 
\label{fig:2}
\end{figure*}

\section{Related Work}
\subsection{Retinex-based LLIE Methods}
The theory of the retinal-cortex (Retinex) is based on the model of color invariance and the subjective perception of color by the human visual system (HVS)~\cite{land1971lightness}. It decomposes the image into illumination and reflectance maps. It has been widely used in low-light image enhancement and has been proven to be effective and reliable.

\textbf{Traditional Approaches}. In some methods, the resolution of illumination and reflectance patterns is realized by a Gaussian filter or a group of filter banks, such as SSR~\cite{jobson1997properties} and MSR~\cite{jobson1997multiscale}. LIME~\cite{guo2016lime} estimates the lighting map by initializing the maximum values of three channels and applying a structural prior refinement to form the final illumination map. JED~\cite{ren2018joint} enhances the image and suppresses noise by combining sequence decomposition and gamma transform. Traditional methods mainly show poor generalization and poor robustness, limiting their application.

\textbf{Deep Learning-based Approaches}. Retinex-Net~\cite{wei2018deep} combines the paradigm of Retinex decomposition with deep learning. It applies a phased decomposition and adjustment structure and uses BM3D~\cite{danielyan2011bm3d} for image denoising. Similarly, KinD~\cite{zhang2019kindling} and KinD++~\cite{zhang2021beyond} adopt the decomposition and adjustment paradigm and use the convolutional neural network (CNN) to learn the mapping in both decomposition and adjustment. Robust Retinex~\cite{zhu2020zero} decomposes the image into three components, \textit{i.e.}, illumination, reflectance, and noise. Then, it estimates the noise and restores the illumination by iterating with the guidance of loss, so as to achieve the purpose of denoising and enhancement. Although these methods show excellent performances, the CNN-based decomposition cannot make full use of global information due to the limitations of convolution. Moreover, they also suffer from some thorny problems, such as the difficulty of designing loss functions and the challenge of completing some missing scene contents.

\subsection{Generative LLIE Methods}
With the development of variational auto encoder (VAE)~\cite{kingma2013auto}, GAN~\cite{wang2017generative, zhu2018generative}, and other generative models, image generation can achieve excellent results. From a new perspective, the generative model can take the low-light image as the condition and generate the corresponding normal-light image to objectively realize the goal of low-light image enhancement. EnlightenGAN~\cite{jiang2021enlightengan} designs a single generative to directly map a low-light image to a normal-light image. It is combined with global and local discriminators to realize the function. CIGAN~\cite{ni2022cycle} uses a cycle-interactive GAN to complete the cycle generation and information transmission of light between normal-light and low-light images. These methods have achieved  results. However, the training process of GAN is difficult and the convergence of the loss function is unstable. Recently, diffusion models~\cite{ho2020denoising, niu2023cdpmsr, ren2022image} have emerged as a powerful family of generative models with record-breaking performance in many domains, including image generation, inpainting, \textit{etc}. It overcomes some shortcomings of GAN and breaks the long-term dominance of GAN in image generation. In this paper, we explore a novel approach to combine Retinex model with the diffusion model for the first time.

\section{Methodology}
The overall framework of Diff-Retinex is summarized as Fig.~\ref{fig:2}. A general Retinex-based enhancement framework should be able to decompose images flexibly and remove various degradations adaptively. Therefore, Transformer decomposition network first decomposes the image into illumination and reflectance maps according to the Retinex theory. Then, the illumination and reflectance maps are adjusted through the multi-path diffusion generation adjustment network (including reflectance diffusion adjustment and illumination diffusion adjustment). The enhanced result is the production of the adjusted components.

\subsection{Transformer Decomposition Network}
The classical Retinex theory assumes that an image can be decomposed into reflectance and illumination maps as:
\begin{equation}
\label{eq1}
I=R\cdot L,
\end{equation}
where $I$ is the input image. $R$ and $L$ denote the reflectance and illumination maps, respectively. It is essentially an ill-posed problem. The reflectance map reflects the scene content, so it tends to be constant in different lighting conditions. The illumination map is related to the lighting condition and should present local smoothness.

Specially, some degraded images may also carry complex noise with varying degrees of pollution. In this condition, we tend to follow the decomposition property that the illumination map is locally smooth. Thus, the noise is decomposed into the reflectance map. The optimization objective to realize Retinex decomposition in our method is generally represented via Eq.~\eqref{eq2}:
\begin{equation}
\label{eq2}
\mathop{\text{min}}\limits_{R,L} \tau(R\cdot L) + \alpha \phi(R) + \beta \psi(L),  
\end{equation}
where $\tau(R\cdot L)$ ensures that the image can be reconstructed from the decomposed illumination and reflectance maps. $\phi(R)$ constrains the consistency of reflectance map. $\psi(L)$ makes the illumination map simple in structure and smooth in segments. $\alpha$ and $\beta$ are the hyper-parameters. The detailed designs of the loss functions are as follows.

\subsubsection{Loss Functions}
Based on Eq.~\eqref{eq2}, we design the following loss functions, including the reconstruction loss, reflectance consistency loss, and illumination smoothness loss, to optimize the Transformer Decomposition Network. Considering the reflectance consistency in different illumination conditions, we use paired low-light and normal-light images for training, denoted as $I_l$ and $I_n$, respectively. The reflectance maps decomposed from them are denoted as $R_l$ and $R_n$, respectively. The corresponding illumination maps are represented by $L_l$ and $L_n$.

\noindent \textbf{Reconstruction Loss $\tau(R\cdot L)$}. It guarantees that the decomposed $R$ and $L$ can reconstruct the original image. Thus, this loss is denoted by considering the image fidelity:
\begin{equation}
\label{eqLrec}
L_{rec}= \|R_n\cdot L_n - I_n\|_{1} + \alpha_{rec}\|R_l\cdot L_l - I_l\|_{1} + \xi(L_{crs}),
\end{equation}
where $\alpha_{rec}$ is the hyper-parameters, applied to adjust the contribution of different illumination. $\xi(L_{crs})$ is an small auxiliary function for the cross multiplication of the illumination and reflection maps for low and normal light.

\noindent \textbf{Reflectance Consistency Loss $\phi(R)$}. Considering that the reflectance of objects is invariant in various lighting conditions, we constrain the consistency of reflectance maps in different lighting conditions. Specifically, It can be described as:
\begin{equation}
\label{eqLrc}
L_{rc}=\|R_{n}-R_{l}\|_{1}.
\end{equation}

\noindent \textbf{Illumination Smoothness Loss $\psi(L)$}. Considering the illumination should be piece-wise smooth, we constrain it by:
\begin{equation}
\label{eqLrc}
L_{smooth}=\|W_{T}^{l}\cdot \nabla L_l\| + \|W_{T}^{n}\cdot \nabla L_n\|,
\end{equation}
where $W_{T}^{l}$ and $W_{T}^{n}$ are the weighting factors. It can be expressed in fractional form or exponential form. To simplify the process, we set $W_{T}^{l}\xleftarrow{} e^{-c\cdot \nabla I_{l}}$ and $W_{T}^{n}\xleftarrow{} e^{-c\cdot \nabla I_{n}}$. $c$ is the constraint factor. $\nabla$ denotes the derivative filter. This loss guarantees that a large penalty is imposed in regions where the image is smooth and the constraint is relaxed in regions where the image illumination is abrupt.

Ultimately, the overall decomposition loss is denoted as:
\begin{equation}
\label{eq6}
L = L_{rec} + \gamma_{rc}L_{rc} + \gamma_{sm}L_{smooth},
\end{equation}
where $\gamma_{rc}$ and $\gamma_{sm}$ are the hyper-parameters.

\subsubsection{Network Architecture} \label{sec_init_illu}
As shown in Fig.~\ref{fig:2}, Transformer decomposition network (TDN) consists of two branches, \textit{i.e.}, the reflectance decomposition branch and the illumination decomposition branch.

Given an image $I\in \mathbb{R}^{H\times W\times 3}$ to be decomposed, TDN firstly obtains its embedding features $F_{init}\in \mathbb{R}^{H\times W\times C}$ through convolution projection. In the illumination decomposition branch, it makes up of several convolutional layers to reduce the amount of calculation on the premise of ensuring the decomposition effect. To ensure the intrinsic characteristics of illumination and reflectance maps, and improve the recovery performance and information retention in the reflectance map, the reflectance decomposition branch is composed of a multi-stage Transformer encoder and decoder. To be specific, the Transformer encoder and decoder are composed of an attention ($Atten$) module and a feed-forward network ($FFN$) module. In general, we denote the computation in the TDN block as:
\begin{equation}
\label{eq_AT}
\hat{F_{i}}=Atten(Norm(F_{i-1}))+F_{i-1},
\end{equation}
\begin{equation}
\label{eq_FFN}
F_{i}=FFN(Norm(\hat{F}_{i}))+\hat{F}_{i},
\end{equation}
where $Norm$ denotes normalization. $F_{i-1}$ represents the input feature map of the current TDN block.

\begin{figure}[t]
\centering
\includegraphics[width=1\linewidth]{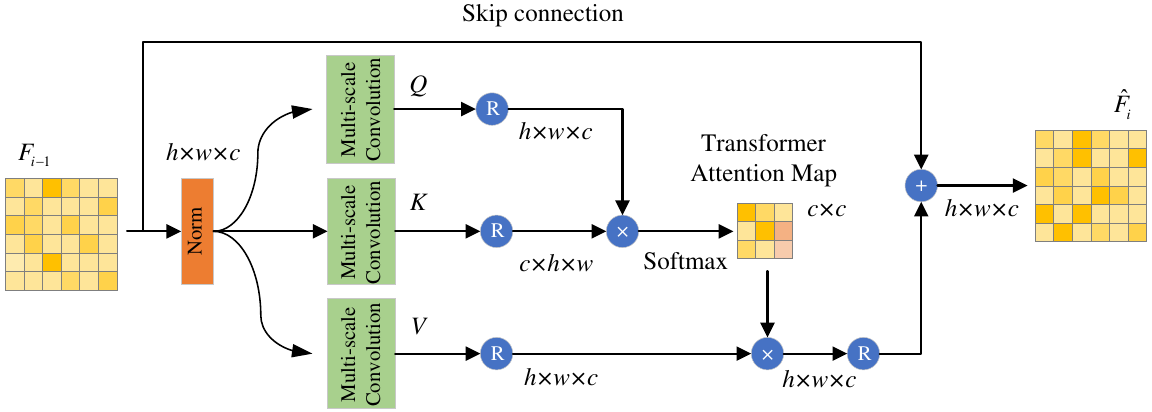}
   \caption{Detailed network architecture of MDLA. The attention is calculated in the direction of cross channel to realize the efficient decomposition of a high-resolution image.}
\label{fig_MDLA}
\end{figure}

Considering the high attention computing overhead in Transformer, the time complexity is proportional to the quadratic size of the image. Therefore, it is not suitable for high-resolution image decomposition. To solve this problem, we design a novel multi-head depth-wise convolutions layer attention (MDLA) for computing attention form in TDN, as shown in Fig.~\ref{fig_MDLA}. On the premise of maintaining the decomposition performance, it reduces the attention computation complexity to a great extent.

In MDLA, for a feature $X\in \mathbb{R}^{h\times w \times c}$ obtained from Layer-Norm, we first aggregate the information of its channel directions with a $1\times 1$ convolution. Subsequently, $3\times3$, $5\times5$ and $7\times7$ convolutions aggregate the information. The outputs of multiple convolutions are queue $Q=W^{qi}_{pc}W^{q}_{dc}X$, key $K=W^{ki}_{pc}W^{k}_{dc}X$, and value $V=W^{vi}_{pc}W^{v}_{dc}X$. We reduce the features dimension by $1\times 1$ convolution, also reshape the features and compute the attention in the direction of the layers. Specifically, it can be formulated as Eq.~\eqref{eq4}:
\begin{equation}
\label{eq4}
\hat{X} = softmax(Q_{R}K_{R}/d)\cdot V_{R} + X,
\end{equation}
where $Q_{R}, V_{R}\in\mathbb{R}^{h\times w\times c}$ and $K_{R}\in \mathbb{R}^{c\times h \times w}$ are $Q$, $V$ and $K$ after reshaping. $d$ is a scale factor.

We take a simple but effective depth-wise separable feed-forward network. It mainly consists of separable point-wise convolution and depth-wise convolution to minimize the amount of computation. Given a feature $X\in \mathbb{R}^{h\times w\times c}$ after Layer-Norm, the output feature can be expressed as:
\begin{equation}
\label{eq5}
\hat{X} = W_{dc}(\phi(W_{pc}W_{dc}(X)))+X,
\end{equation}
where $W_{pc}$ and $W_{dc}$ are the point-wise and depth-wise convolutions. $\phi$ is the activation function.

\subsection{Diffusion Generation Adjustment}
The diffusion generation adjustment aims to construct the original data distribution of Retinex model that recovers multiple channels. Generally, it can be divided into two paths, namely reflectance diffusion adjustment (RDA) and illumination diffusion adjustment (IDA).

\begin{figure}[t]
\centering
\includegraphics[width=1\linewidth]{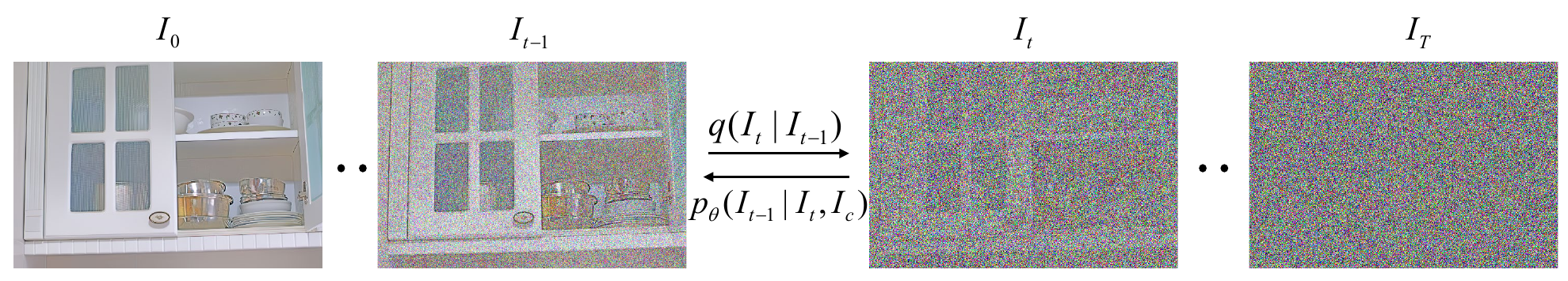}
   \caption{Example of the forward and reverse diffusion processes of Diffusion Generation Adjustment. $I_{0}$ is the obtained result.}
\label{fig:4}
\end{figure}

The normal-light image component is denoted as $I_{0}\in \mathbb{R}^{H\times W \times C}$ ($C=3$ in RDA, $C=1$ in IDA) for diffusion. The conditional images are respectively concatenated with the noisy image to form the guidance. We adopt the diffusion process proposed in Denoising Diffusion Probabilistic Model (DDPM)~\cite{ho2020denoising} to construct the distribution of Retinex data for each channel. More specifically, it can be described as a forward diffusion process and a reverse diffusion process, as shown in Fig.~\ref{fig:4}.

\textbf{Forward Diffusion Process}. 
The forward diffusion process can be viewed as a Markov chain progressively adding Gaussian noise to the data. The data at step $t$ is only dependent on that at step $t-1$. Thus, at any $t\in [0,T]$, we can obtain the data distribution of the noisy image $I_{t}$ as:
\begin{equation}
\label{eq7}
q(I_{t}|I_{t-1})=\mathcal{N}(I_{t}; \sqrt{1-\beta_{t}}I_{t-1}, \beta_{t}\mathcal{Z}),
\end{equation}
where $\beta_{t}$ is a variable controlling the variance of the noise added to the data. When $\beta_{t}$ is small enough, $I_{t-1}$ to $I_{t}$ is a constant process of adding a small amount of noise, \textit{i.e.}, the distribution at step $t$ is equal to that at the previous step adding Gaussian noise. By introducing a new variable $\alpha_{t} = 1 - \beta_{t}$, this process can be described as:
\begin{equation}
\label{eq8}
I_{t} = \sqrt{\alpha_{t}}I_{t-1}+\sqrt{1-\alpha_{t}} \epsilon_{t-1}, \quad \epsilon_{t-1}\sim \mathcal{N}(0,\mathcal{Z}).
\end{equation}

With parameter renormalization, multiple Gaussian distributions are merged and simplified. We can obtain the distribution of the $t$-th step $q(I_{t}|I_{0})$. More specifically, it can be expressed as:
\begin{equation}
\label{eq9}
q(I_{t}|I_{0})=\mathcal{N}(I_{t}; \sqrt{\overline{\alpha}_{t}}I_{0}, (1-\overline{\alpha}_{t})\mathcal{Z}),
\end{equation}
where $\overline{\alpha}_{t}=\prod_{i=0}^{t} \alpha_{i}$. When the distribution $q(I_{t}|I_{0})$ approaches $\mathcal{N}(0,\mathcal{Z})$, the model can be considered to complete the forward process of diffusion.

\textbf{Reverse Diffusion Process}. The reverse diffusion process is the process of restoring the original distribution from the Gaussian distribution of pure noise. Similar to the forward diffusion process, the denoising diffusion process is also carried out in steps. At step $t$, a denoising operation is applied to data $I_{t}$ to get the probability distribution of $I_{t-1}$ with the guidance of the conditional image $I_{c}$. Therefore, given $I_{t}$, we can formulate the conditional probability distribution $I_{t-1}$ as:
\begin{equation}
\label{eq10}
p_{\theta}(I_{t-1}|I_{t}, I_{c})=\mathcal{N}(I_{t-1}; \mu_{\theta}(I_{t}, I_{c}, t), \sigma_{t}^{2}\mathcal{Z}),
\end{equation}
where $\mu_{\theta}(I_{t}, I_{c}, t)$ is the mean value, from the estimate of step $t$.  $\sigma_{t}^{2}$ is the variance. In RDA and IDA, we follow the setup of DDPM and set it to a fixed value. In more detail, they can be further expressed as:
\begin{equation}
\label{eq11}
\mu_{\theta}(I_{t}, I_{c}, t)=\frac{1}{\sqrt{\alpha}_{t}}(I_{t}-\frac{\beta_{t}}{(1-\overline{\alpha}_{t})}\epsilon_{\theta}(I_{t}, I_{c}, t)),
\end{equation}
\begin{equation}
\label{eq12}
\sigma_{t}^{2}=\frac{1-\overline{\alpha}_{t-1}}{1-\overline{\alpha}_{t}}\beta_{t},
\end{equation}
where $\epsilon_{\theta}(I_{t}, I_{c}, t)$ is the estimated value with a deep neural network, given the input $I_{t}$, $I_{c}$ and the time step $t$.

In each step of the reverse diffusion process $t\in [0, T]$, we optimize an objective function for the noise estimated by the network and the noise $\epsilon$ actually added. Therefore, the loss function of the reverse diffusion process is:
\begin{equation}
\label{eq13}
L_{diff}(\theta)=\|\epsilon-\epsilon_{\theta}(\sqrt{\overline{\alpha}_{t}} I_{0}+\sqrt{1-\overline{\alpha}_{t}}\epsilon, I_{c}, t)\|.
\end{equation}

The denoising network in the reverse diffusion process usually combines the characteristics of UNet and attention. In RDA and IDA, we adopt the backbone of SR3~\cite{saharia2022image} and follow the designs of the diffusion denoising network, consisting of multiple stacked residual blocks combined with attention. From the noise predicted by the network, we can estimate the approximate $\widetilde{I}_{0}$. It makes sense to keep the approaching $\widetilde{I}_{0}$ and the normal-light image with consistent content information. We adopt a consistent network to implement this process:
\begin{equation}
\label{eq14}
\widetilde{I}_{0}=\frac{1}{\sqrt{\overline{\alpha}_{t}}}(I_{t}-\sqrt{1-\overline{\alpha}_{t}}\epsilon_{\theta}(I_{t}, I_{c}, t)),
\end{equation}
\begin{equation}
\label{eq15}
L_{content}=\|I_{0}-\epsilon_{c}(\widetilde{I}_{0}, t)\|_{1}.
\end{equation}

In the consistent network $\epsilon_{c}$, the RDA part adopts the backbone of Restormer~\cite{zamir2022restormer} and adds feature affine with the time embedding. The IDA part adopts the structure same as the denoising network. The loss function of the overall diffusion model network is given by:
\begin{equation}
\label{eq16}
L=L_{diff}+\gamma_{ct} L_{content}.
\end{equation}

\begin{figure*}[t]
\centering
\includegraphics[width=0.975\linewidth]{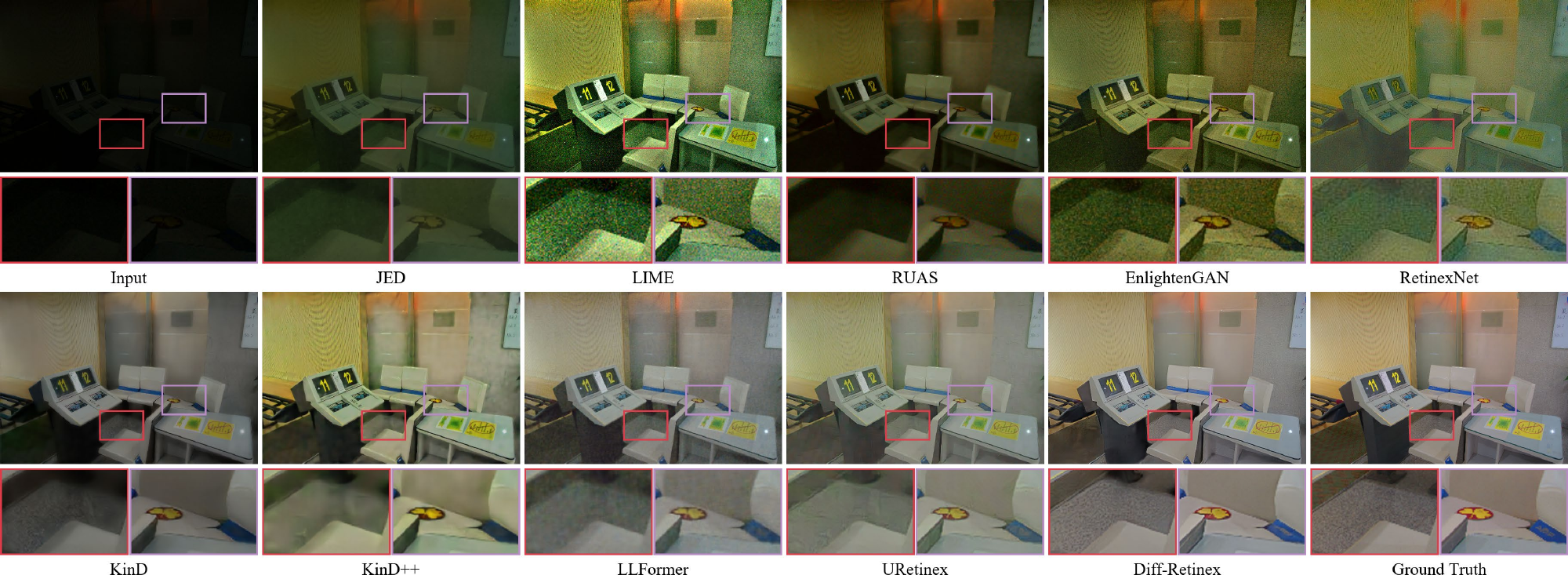}
\vspace{-0.02in}
\caption{Qualitative comparison with the state-of-the-art low-light image enhancement methods on the LOL dataset.}
\label{fig_cmp1}
\end{figure*}

In general, the whole diffusion generation adjustment process is to restore the original Retinex decomposition distribution from the low-light Retinex decomposition distribution. We can formulate the whole diffusion process as:
\begin{equation}
\label{eq14}
\hat{R}_{DGA}=\mathcal{F}_{RDA}(\epsilon_{s}^{(r)},R_{TDN}),
\end{equation}
\begin{equation}
\label{eq15}
\hat{L}_{DGA}=\mathcal{F}_{IDA}(\epsilon_{s}^{(i)},L_{TDN}),
\end{equation}
where $\epsilon_{s}^{(r)}\in \mathbb{R}^{H\times W\times 3}$ and $\epsilon_{s}^{(i)}\in \mathbb{R}^{H\times W\times 1}$ are Gaussian noise generated by initialization. $R_{TDN}$ and $L_{TDN}$ are the reflectance and illumination maps obtained by TDN.

Ultimately, the enhanced image is obtained as the production of the diffusion generative adjusted illumination and reflectance maps, \textit{i.e.}, $\hat{I}=\hat{R}_{DGA}\cdot \hat{L}_{DGA}$.

\section{Experiment}
\subsection{Implementation Details and Datasets}
\textbf{Implementation Details.} The proposed Diff-Retinex is separately trained. TDN is first trained. Empirically, we set $\gamma_{rc}=0.1$, $\gamma_{sm}=0.1$, $\alpha_{rec}=0.3$. The learning rate is $lr=0.0001$ and the batch size is 16 with Adam optimizer. Then, we train the networks related to diffusion generation adjustment. The steps of IDA and RDA are set to $t=1000$, $\gamma_{ct}=1$ The input image is of size $160\times 160$, the batch size is 16. Adam optimizer with a learning rate of 0.0001 is used to train 800K iterations on the network. All the experiments are conducted on the NVIDIA GeForce RTX 3090 GPU with PyTorch~\cite{paszke2019pytorch} framework.

\textbf{Datasets.} To verify the generalization, we conduct experiments on the LOL~\cite{wei2018deep} and VE-LOL-L~\cite{liu2021benchmarking} datasets. All the images in the LOL datasets are taken in real life. We use 485 pairs of images for training and 15 low-light images for testing. The VE-LOL dataset contains data for high-level and low-level visual tasks, called VE-LOL-H and VE-LOL-L, respectively. VE-LOL-L is also adopted to evaluate the effectiveness of our method. DICM is used as the evaluation dataset for generalization by cross-testing.

\begin{figure*}[h]
\centering
\includegraphics[width=0.98\linewidth]{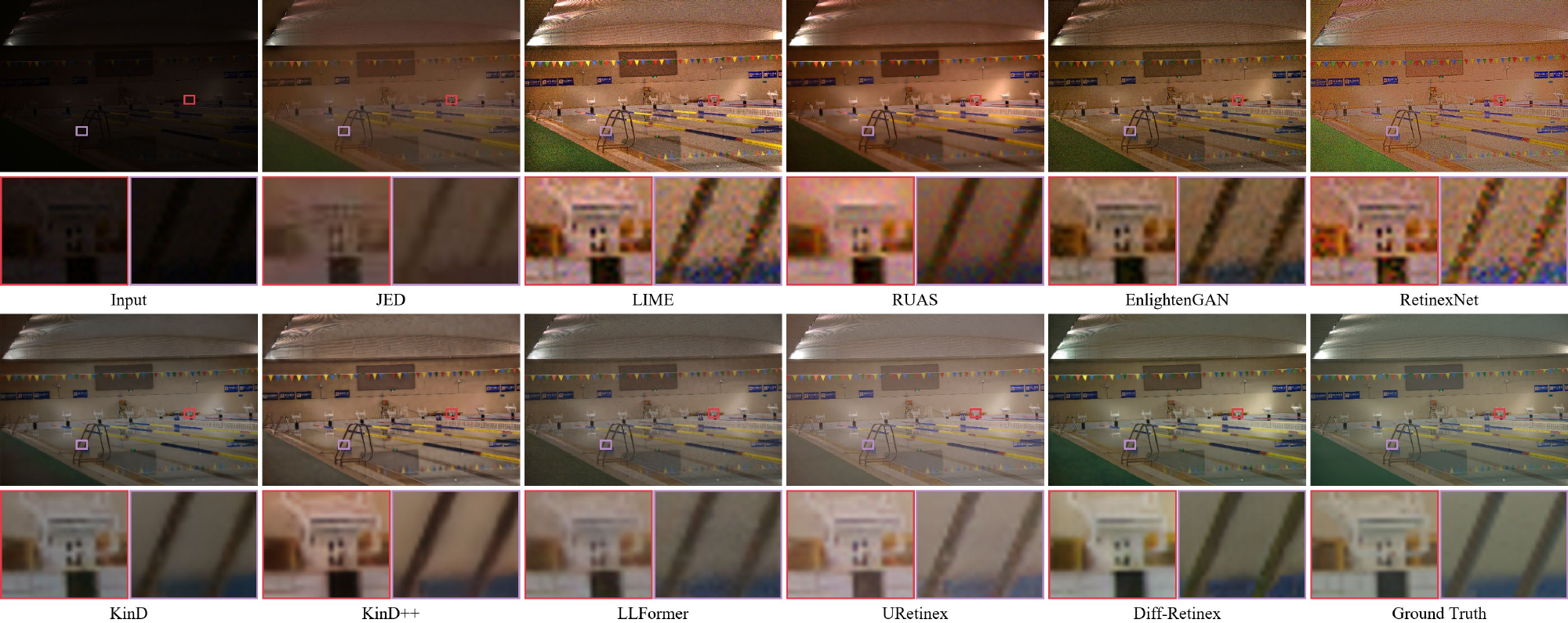}
   \caption{Qualitative comparison with the state-of-the-art low-light image enhancement methods on the VE-LOL-L dataset.}
\label{fig_cmp2}
\end{figure*}

\begingroup
\setlength{\tabcolsep}{6pt}{
\renewcommand{\arraystretch}{1.2} 
\begin{table*}[t]
	\centering
	\caption{\label{tab:cmp} Quantitative results of low-light image enhancement methods on the LOL and VE-LOL-L datasets.}
\vspace{0.05in}
  \resizebox{\textwidth}{!}
{
	\begin{tabu}{l|c|cccccccccccc}
		\tabucline[1.5pt]{-}
		 &Methods & EnlightenGAN & JED & Robust Retinex & RUAS & KinD & KinD++ & LIME &RetinexNet& Zero-DCE & URetinex & LLFormer & Diff-Retinex\\
		\hline
		\multirow{4}{*}{\rotatebox{90}{LOL}} &FID$\downarrow$ & 105.59 & 105.86 & 92.32 & 95.59 & 78.59 & 110.68 & 114.00 & 150.50 & 106.63 & \underline{59.00} & 76.96 & \textbf{47.85}\\
  		&LPIPS$\downarrow$ & 0.129 & 0.190 & 0.157 & 0.167  & 0.083 & 0.095 & 0.211 & 0.183 & 0.133 & \underline{0.050} & 0.067 & \textbf{0.048}\\
        &BIQI$\downarrow$  & 30.95 & 32.33 & 42.29 & 43.32 & 26.70 & 26.81 & 37.83 & 29.33 & 34.80 & \underline{23.05} & 28.81 & \textbf{19.97}\\
        &LOE$\downarrow$	& 395.52 & 306.90 & 202.31 & 195.36 &758.56 &700.79 &547.54 &395.33 &232.97 &197.02&\textbf{176.61} &\underline{191.56}\\
        \tabucline[1.5pt]{-}
		\hline
		\multirow{4}{*}{\rotatebox{90}{VE-LOL-L}} &FID$\downarrow$ & 92.58 & 110.46 & 79.64 & 100.07  & 65.56 & 98.10 & 98.90 & 158.99 & 93.81 & \underline{48.36} & 79.83 & \textbf{47.75}\\
  		&LPIPS$\downarrow$ & 0.124 & 0.158 & 0.106 & 0.144  & \underline{0.070} & 0.114 & 0.248 & 0.283 & 0.123 & 0.091 & 0.110 & \textbf{0.050}\\
        &BIQI$\downarrow$  & 32.77 & \underline{27.29} & 39.75 & 32.51 & 28.23 & 32.33 & 47.09 & 45.59 &35.06& 35.39 & 32.47 &\textbf{26.54}\\
        &LOE$\downarrow$   & 422.77 & 330.25 & \textbf{128.73} &168.99 &239.33 &623.63 &554.69 &531.92 &228.88 & 166.02 & 177.87 &\underline{149.60}\\
        \tabucline[1.5pt]{-}
	\end{tabu}
 }
\end{table*}
}

\subsection{Results and Analysis}
We present quantitative and qualitative comparisons with state-of-the-art methods including both traditional and deep learning-based methods. The traditional methods include LIME~\cite{guo2016lime} based on illumination estimation and JED~\cite{ren2018joint} based on Retinex decomposition and joint denoising. The learning-based methods include RetinexNet~\cite{wei2018deep}, KinD~\cite{zhang2019kindling}, KinD++~\cite{zhang2021beyond}, RUAS~\cite{liu2021retinex}, EnlightenGAN~\cite{jiang2021enlightengan}, URetinex~\cite{wu2022uretinex}, and LLFormer~\cite{wang2023ultra}.

\textbf{Qualitative Comparison}. Qualitative results are shown in Figs.~\ref{fig_cmp1} and \ref{fig_cmp2}. Our method shows three obvious advantages. \textit{First and foremost, Diff-Retinex has the ability of texture completion and reasoning generation for missing scenes}. It is a remarkable characteristic of our generative diffusion model and not possessed in existing methods. As shown in Fig.~\ref{fig_cmp1}, the highlighted region on the right is ground with coarse-grained textured tile (see ground truth). All the competitors fail to recover the coarse-grained texture tile while our method can generate the missing textures similar to the ground truth. Similarly, the diving platform and handrail in Fig.~\ref{fig_cmp2} are seriously missing and damaged in the low-light image. Most methods cannot complete the clear textures while Diff-Retinex can. \textit{Second, our method shows better illumination and color fidelity.} In Fig.~\ref{fig_cmp1}, the low-light image has considerable color deviation. In the whole view, the color of Diff-Retinex is the closest to that of the ground truth. KinD, KinD++, RetinexNet, and URetinex appear with different degrees of color deviation, \textit{e.g.}, URetinex and KinD++ tend to be yellow. In Fig.~\ref{fig_cmp2}, Diff-Retinex also performs better than other SOTA methods in the color of the venue. \textit{Last, our results exhibit vivid textures with less noise than other methods.} LIME and RetinexNet have much noise left in the whole image, affecting the scene expression. The denoising performance of EnlightenGAN and LLFormer at the flat place is unsatisfactory, \textit{e.g.}, the black area below the computer desk and wall in Fig.~\ref{fig_cmp1}. In general, Diff-Retinex shows obvious advantages in these areas.

\begingroup
\setlength{\tabcolsep}{10pt}
\renewcommand{\arraystretch}{1.2}
\begin{table}[t]
\caption{\label{tab:PSNR} Quantitative comparison of PSNR and SSIM on LOL.}
\vspace{0.03in}
\footnotesize
	\centering
 \resizebox{0.45\textwidth}{!}{
	\begin{tabu}{c|c|cc}
		\tabucline[0.8pt]{-}
		 Method & Main Type & PSNR$\uparrow$ & SSIM$\uparrow$\\
		\hline
		  RetinexNet & \textit{CNN} & 17.56 & 0.698 \\
            KinD & \textit{CNN} & 17.64 & 0.829 \\
            KinD++ & \textit{CNN} & 17.75 & 0.816 \\
            EnlightenGAN & \textit{GAN} & 17.48 & 0.716 \\
            URetinex & \textit{Unfolding} & 21.32 & 0.836  \\
            LLFormer & \textit{Transformer} & \textbf{23.66} & \textbf{0.873} \\
            Diff-Retinex & \textit{Diffusion} & \underline{21.98} & \underline{0.863} \\
		\tabucline[1pt]{-}
	\end{tabu}}
\end{table}

\textbf{Quantitative Comparison}. Metrics including FID~\cite{heusel2017gans}, LPIPS~\cite{zhang2018unreasonable}, BIQI~\cite{moorthy2010two}, LOE~\cite{wang2013naturalness} and PI~\cite{blau20182018} are adopted for evaluation. FID is the machine feature similarity used to evaluate image similarity. LPIPS is learned perceptual image patch similarity, which measures the image differences. BIQI is an image-blind quality evaluation index. LOE is the sequence error of image brightness, reflecting the natural retention ability of the image. PI represents the subjective perceived quality of the image. The lower FID, LPIPS, BIQI, LOE, and PI, the better the image quality. The quantitative results on LOL and VE-LOL-L datasets are reported in Tab.~\ref{tab:cmp}. For LOL, our method shows great advantages over other methods in generative indicators FID and LPIPS. It indicates that our results have better generation similarity of machine vision. In terms of brightness sequence error, our method is slightly lower than LLFormer. However, benefit from generative diffusion model and TDN, it shows the best performance among all the Retinex-based methods, including RetinexNet, KinD, KinD++, and URetinex. For VE-LOL-L, our method achieves the comprehensive best performance from the perspective of metrics as well. It indicates that our method has strong generalization and advanced generation enhancement performance in various scenarios. For DICM, in Fig.~\ref{fig_TDN_abc}, our method also demonstrates competitiveness. In addition, we also provide the quantitative comparisons of PSNR and SSIM in Tab.~\ref{tab:PSNR}.

\begin{figure}[t]
	\centering
	\includegraphics[width=0.96\linewidth]{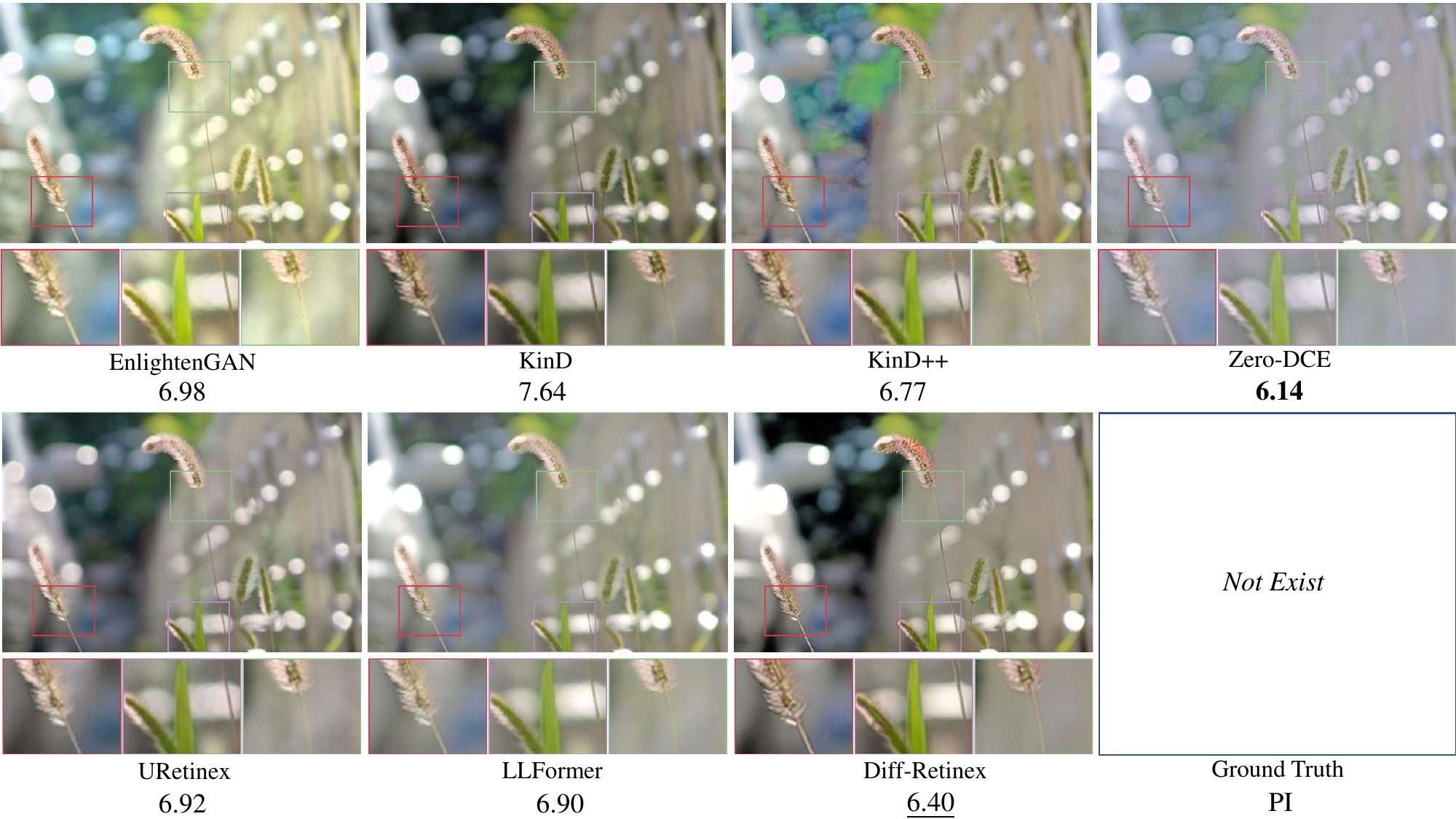}
    \vspace{-0.03in}
	\caption{Qualitative and quantitative comparison of PI with the SOTA low-light image enhancement methods on DICM.}
	\label{fig_TDN_abc}
\end{figure}

\subsection{Ablation Study}
\textbf{Transformer Decomposition Network}. To validate the effectiveness of Transformer decomposition network (TDN), we visualize the decomposition. Retinex decomposition is an ill-posed problem with no exact optimal solution. A core point is that the reflectance information should be strictly consistent in different illumination levels. Typical and effective representation methods adopt CNN for decomposition, such as RetinexNet and KinD++. The reflectance decomposition results are shown in Fig.~\ref{fig_TDN_ab}.

\begin{figure}[t]
\centering
\includegraphics[width=0.98\linewidth]{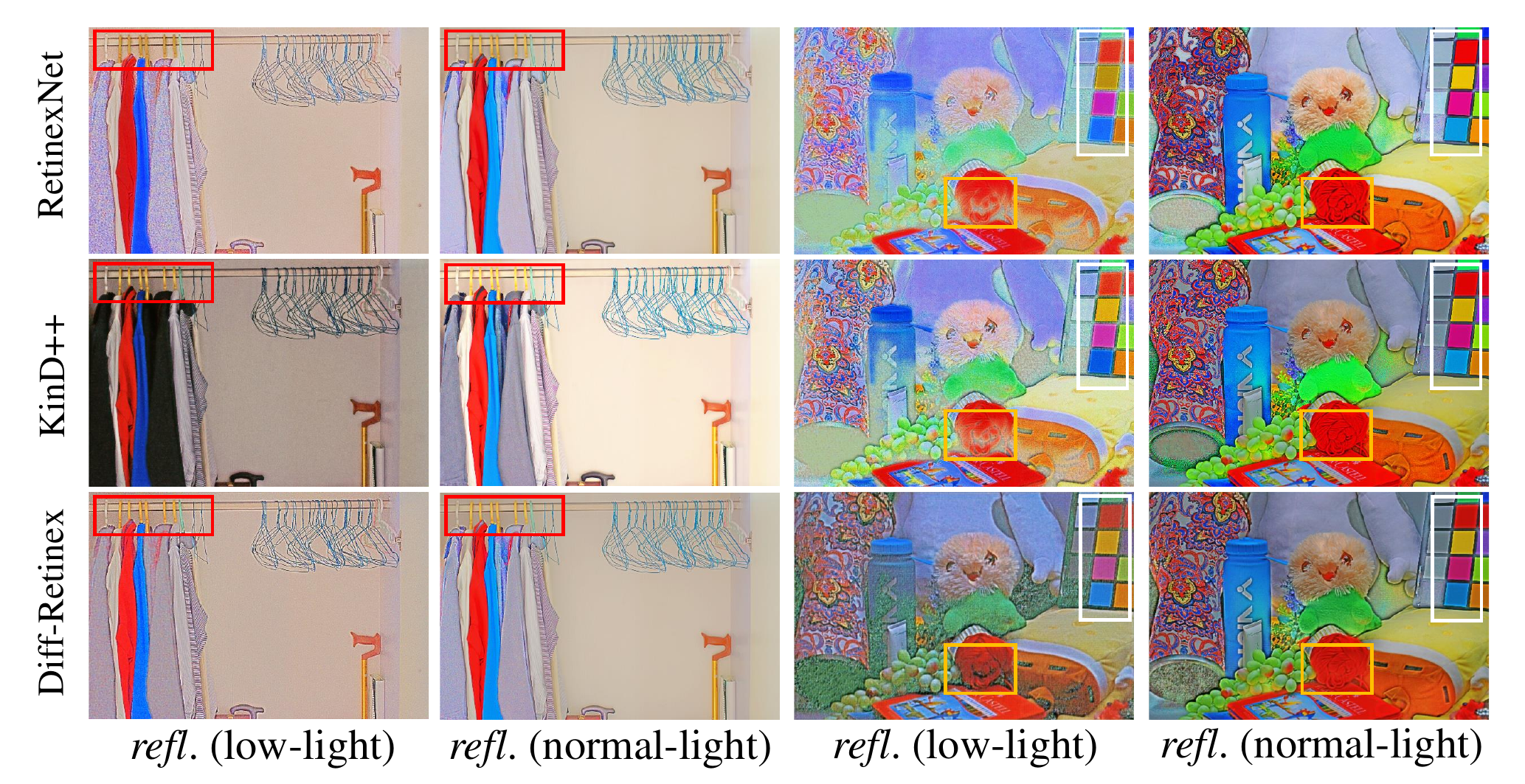}
   \caption{Qualitative comparison on reflection map of the decomposition network. The structure details and noise are assumed to be decomposed into the reflection map.}
\label{fig_TDN_ab}
\end{figure}

\begin{figure}[t]
\centering
\includegraphics[width=0.98\linewidth]{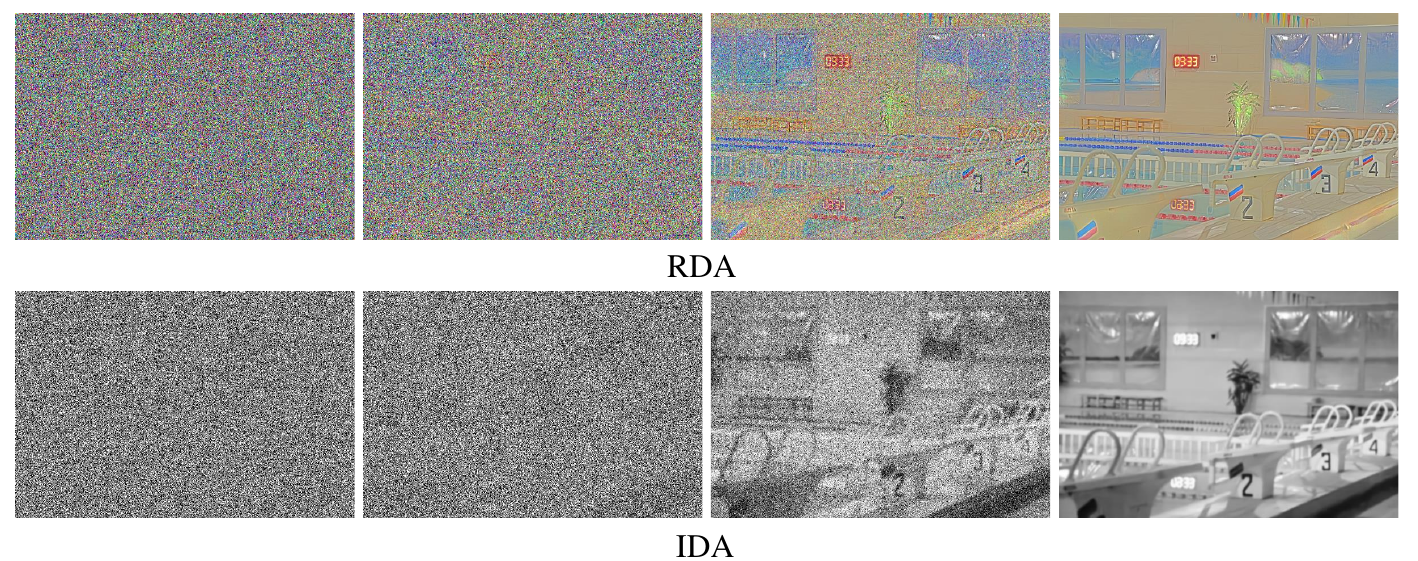}
\caption{Example of RDA and IDA. To better show the diffusion effect, the output format of constant period sampling is adopted. Left to right: iterative process of gradual recovery from pure noise.}
\label{fig_RDA_IDA}
\end{figure}

\textbf{Generative Diffusion Model.} To validate the effectiveness of the diffusion model, on the one hand, we visualize the generation process of the RDA and IDA, as shown in Fig.~\ref{fig_RDA_IDA}. On the other hand, we compare the restoration results of the reflectance map through our diffusion model and some other one-step Retinex-based LLIE methods. We use the reflectance map to compare for it contains a lot of color and texture information which is more sensitive to visual perception. Typical Retinex-based LLIE methods include RetinexNet and KinD++. For reflectance restoration, RetinexNet adopts BM3D and KinD++ adopts CNN. The results are shown in Fig.~\ref{fig_difab}. For Retinex decomposition results of methods are quiet different, we show their own decomposed reflectance map from the normal-light image as ground truth for comparison. It can be seen that our method can better deal with color deviation and shows better performance on texture restoration. We also calculate the FID, LPIPS, and BIQI between the restored reflectance map and corresponding ground truth for quantitative evaluation. The results are reported in Tab.~\ref{tab3}.

\begin{figure}[t]
\centering
\includegraphics[width=1\linewidth]{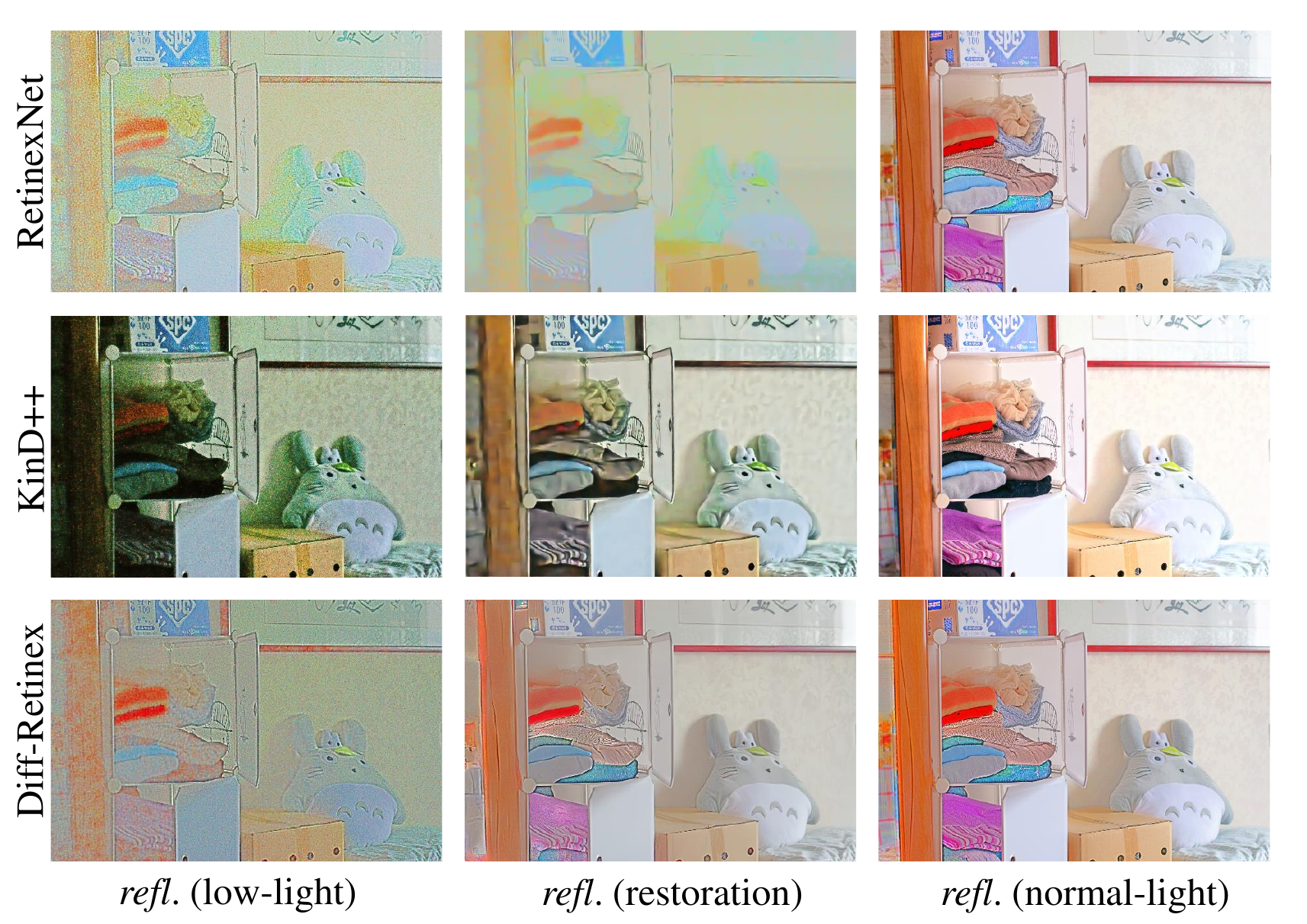}
\caption{Qualitative comparison of the restoration of reflectance map with our diffusion model and other SOTA methods.}
\label{fig_difab}
\label{fig:10}
\end{figure}

\begingroup
\setlength{\tabcolsep}{10pt}
\renewcommand{\arraystretch}{1.2}
\begin{table}[t]
\caption{\label{tab3} Quantitative comparison on reflectance restoration with the diffusion model and other low-light enhancement methods.}
\vspace{0.05in}
\footnotesize
	\centering
	\begin{tabu}{c|ccc}
		\tabucline[0.8pt]{-}
		 Method & FID$\downarrow$ & LPIPS$\downarrow$ & BIQI$\downarrow$\\
		\hline
		 RetinexNet (\textit{BM3D}) & 111.29 & 0.225 &  24.80 \\
         KinD++ (\textit{CNN}) & 171.45 & 0.110 &  36.99 \\
         Diff-Retinex (\textit{Diff.}) & \textbf{61.33} & \textbf{0.059} &  \textbf{18.98} \\
		\tabucline[1pt]{-}
	\end{tabu}
\end{table}

\subsection{Discussion}
While Diff-Retinex, functioning as a generative model for low-light image enhancement, exhibits commendable visual outcomes, it does not establish dominance at the pixel-wise error metrics, \textit{e.g.}, PNSR, as shown in Tab.~\ref{tab:PSNR}. Higher PSNR can be attained through more stringent constraints, but the generation effect will be weakened to some extent. In this paper, we encourage the adoption of generative Diffusion to explore the possibility of generative effects for low-light enhancement tasks. Naturally, it is also feasible and desirable to achieve better performance with the pixel-level error through diffusion models.

\section{Conclusion}
In this paper, we rethink the low-light image enhancement task and propose a generative Diff-Retinex model. Diff-Retinex formulates the low-light enhancement task as a paradigm of decomposition and image generation. It can adaptively decompose images into illumination and reflectance maps and solve various degradations by generating diffusion models. The experimental results show that Diff-Retinex has excellent performance and makes subtle-detail completion and inference restoration of low-light image enhancement into reality.


{\small
\bibliographystyle{ieee_fullname}
\bibliography{egbib}

\begin{thebibliography}{10}\itemsep=-1pt

\bibitem{blau20182018}
Yochai Blau, Roey Mechrez, Radu Timofte, Tomer Michaeli, and Lihi Zelnik-Manor.
\newblock The 2018 pirm challenge on perceptual image super-resolution.
\newblock In {\em Proceedings of the European Conference on Computer Vision
  (ECCV) Workshops}, pages 0--0, 2018.

\bibitem{danielyan2011bm3d}
Aram Danielyan, Vladimir Katkovnik, and Karen Egiazarian.
\newblock Bm3d frames and variational image deblurring.
\newblock {\em IEEE Transactions on Image Processing}, 21(4):1715--1728, 2011.

\bibitem{fu2016fusion}
Xueyang Fu, Delu Zeng, Yue Huang, Yinghao Liao, Xinghao Ding, and John Paisley.
\newblock A fusion-based enhancing method for weakly illuminated images.
\newblock {\em Signal Processing}, 129:82--96, 2016.

\bibitem{guo2016lime}
Xiaojie Guo, Yu Li, and Haibin Ling.
\newblock Lime: Low-light image enhancement via illumination map estimation.
\newblock {\em IEEE Transactions on Image Processing}, 26(2):982--993, 2016.

\bibitem{han2010novel}
Ji-Hee Han, Sejung Yang, and Byung-Uk Lee.
\newblock A novel 3-d color histogram equalization method with uniform 1-d gray
  scale histogram.
\newblock {\em IEEE Transactions on Image Processing}, 20(2):506--512, 2010.

\bibitem{heusel2017gans}
Martin Heusel, Hubert Ramsauer, Thomas Unterthiner, Bernhard Nessler, and Sepp
  Hochreiter.
\newblock Gans trained by a two time-scale update rule converge to a local nash
  equilibrium.
\newblock {\em Advances in Neural Information Processing Systems}, 30, 2017.

\bibitem{ho2020denoising}
Jonathan Ho, Ajay Jain, and Pieter Abbeel.
\newblock Denoising diffusion probabilistic models.
\newblock {\em Advances in Neural Information Processing Systems},
  33:6840--6851, 2020.

\bibitem{jiang2021enlightengan}
Yifan Jiang, Xinyu Gong, Ding Liu, Yu Cheng, Chen Fang, Xiaohui Shen, Jianchao
  Yang, Pan Zhou, and Zhangyang Wang.
\newblock Enlightengan: Deep light enhancement without paired supervision.
\newblock {\em IEEE Transactions on Image Processing}, 30:2340--2349, 2021.

\bibitem{jobson1997multiscale}
Daniel~J Jobson, Zia-ur Rahman, and Glenn~A Woodell.
\newblock A multiscale retinex for bridging the gap between color images and
  the human observation of scenes.
\newblock {\em IEEE Transactions on Image Processing}, 6(7):965--976, 1997.

\bibitem{jobson1997properties}
Daniel~J Jobson, Zia-ur Rahman, and Glenn~A Woodell.
\newblock Properties and performance of a center/surround retinex.
\newblock {\em IEEE Transactions on Image Processing}, 6(3):451--462, 1997.

\bibitem{kingma2013auto}
Diederik~P Kingma and Max Welling.
\newblock Auto-encoding variational bayes.
\newblock {\em arXiv preprint arXiv:1312.6114}, 2013.

\bibitem{kong2021low}
Xiang-Yu Kong, Lei Liu, and Yun-Sheng Qian.
\newblock Low-light image enhancement via poisson noise aware retinex model.
\newblock {\em IEEE Signal Processing Letters}, 28:1540--1544, 2021.

\bibitem{land1971lightness}
Edwin~H Land and John~J McCann.
\newblock Lightness and retinex theory.
\newblock {\em Josa}, 61(1):1--11, 1971.

\bibitem{lee2013contrast}
Chulwoo Lee, Chul Lee, and Chang-Su Kim.
\newblock Contrast enhancement based on layered difference representation of 2d
  histograms.
\newblock {\em IEEE Transactions on Image Processing}, 22(12):5372--5384, 2013.

\bibitem{li2018lightennet}
Chongyi Li, Jichang Guo, Fatih Porikli, and Yanwei Pang.
\newblock Lightennet: A convolutional neural network for weakly illuminated
  image enhancement.
\newblock {\em Pattern Recognition Letters}, 104:15--22, 2018.

\bibitem{li2018structure}
Mading Li, Jiaying Liu, Wenhan Yang, Xiaoyan Sun, and Zongming Guo.
\newblock Structure-revealing low-light image enhancement via robust retinex
  model.
\newblock {\em IEEE Transactions on Image Processing}, 27(6):2828--2841, 2018.

\bibitem{liu2021benchmarking}
Jiaying Liu, Dejia Xu, Wenhan Yang, Minhao Fan, and Haofeng Huang.
\newblock Benchmarking low-light image enhancement and beyond.
\newblock {\em International Journal of Computer Vision}, 129:1153--1184, 2021.

\bibitem{liu2022learning}
Risheng Liu, Long Ma, Tengyu Ma, Xin Fan, and Zhongxuan Luo.
\newblock Learning with nested scene modeling and cooperative architecture
  search for low-light vision.
\newblock {\em IEEE Transactions on Pattern Analysis and Machine Intelligence},
  45(5):5953--5969, 2022.

\bibitem{liu2021retinex}
Risheng Liu, Long Ma, Jiaao Zhang, Xin Fan, and Zhongxuan Luo.
\newblock Retinex-inspired unrolling with cooperative prior architecture search
  for low-light image enhancement.
\newblock In {\em Proceedings of the IEEE/CVF Conference on Computer Vision and
  Pattern Recognition}, pages 10561--10570, 2021.

\bibitem{liu2021underexposed}
Risheng Liu, Long Ma, Yuxi Zhang, Xin Fan, and Zhongxuan Luo.
\newblock Underexposed image correction via hybrid priors navigated deep
  propagation.
\newblock {\em IEEE Transactions on Neural Networks and Learning Systems},
  33(8):3425--3436, 2021.

\bibitem{liu2021swin}
Ze Liu, Yutong Lin, Yue Cao, Han Hu, Yixuan Wei, Zheng Zhang, Stephen Lin, and
  Baining Guo.
\newblock Swin transformer: Hierarchical vision transformer using shifted
  windows.
\newblock In {\em Proceedings of the IEEE/CVF International Conference on
  Computer Vision}, pages 10012--10022, 2021.

\bibitem{lore2017llnet}
Kin~Gwn Lore, Adedotun Akintayo, and Soumik Sarkar.
\newblock Llnet: A deep autoencoder approach to natural low-light image
  enhancement.
\newblock {\em Pattern Recognition}, 61:650--662, 2017.

\bibitem{ma2023bilevel}
Long Ma, Dian Jin, Nan An, Jinyuan Liu, Xin Fan, and Risheng Liu.
\newblock Bilevel fast scene adaptation for low-light image enhancement.
\newblock {\em arXiv preprint arXiv:2306.01343}, 2023.

\bibitem{ma2022low}
Long Ma, Risheng Liu, Yiyang Wang, Xin Fan, and Zhongxuan Luo.
\newblock Low-light image enhancement via self-reinforced retinex projection
  model.
\newblock {\em IEEE Transactions on Multimedia}, 2022.

\bibitem{ma2022toward}
Long Ma, Tengyu Ma, Risheng Liu, Xin Fan, and Zhongxuan Luo.
\newblock Toward fast, flexible, and robust low-light image enhancement.
\newblock In {\em Proceedings of the IEEE/CVF Conference on Computer Vision and
  Pattern Recognition}, pages 5637--5646, 2022.

\bibitem{ma2022practical}
Long Ma, Tianjiao Ma, Xinwei Xue, Xin Fan, Zhongxuan Luo, and Risheng Liu.
\newblock Practical exposure correction: Great truths are always simple.
\newblock {\em arXiv preprint arXiv:2212.14245}, 2022.

\bibitem{moorthy2010two}
Anush~Krishna Moorthy and Alan~Conrad Bovik.
\newblock A two-step framework for constructing blind image quality indices.
\newblock {\em IEEE Signal Processing Letters}, 17(5):513--516, 2010.

\bibitem{ni2022cycle}
Zhangkai Ni, Wenhan Yang, Hanli Wang, Shiqi Wang, Lin Ma, and Sam Kwong.
\newblock Cycle-interactive generative adversarial network for robust
  unsupervised low-light enhancement.
\newblock In {\em Proceedings of the ACM International Conference on
  Multimedia}, pages 1484--1492, 2022.

\bibitem{niu2023cdpmsr}
Axi Niu, Kang Zhang, Trung~X Pham, Jinqiu Sun, Yu Zhu, In~So Kweon, and Yanning
  Zhang.
\newblock Cdpmsr: Conditional diffusion probabilistic models for single image
  super-resolution.
\newblock {\em arXiv preprint arXiv:2302.12831}, 2023.

\bibitem{paszke2019pytorch}
Adam Paszke, Sam Gross, Francisco Massa, Adam Lerer, James Bradbury, Gregory
  Chanan, Trevor Killeen, Zeming Lin, Natalia Gimelshein, Luca Antiga, et~al.
\newblock Pytorch: An imperative style, high-performance deep learning library.
\newblock {\em Advances in Neural Information Processing Systems}, 32, 2019.

\bibitem{ren2022image}
Mengwei Ren, Mauricio Delbracio, Hossein Talebi, Guido Gerig, and Peyman
  Milanfar.
\newblock Image deblurring with domain generalizable diffusion models.
\newblock {\em arXiv preprint arXiv:2212.01789}, 2022.

\bibitem{ren2018joint}
Xutong Ren, Mading Li, Wen-Huang Cheng, and Jiaying Liu.
\newblock Joint enhancement and denoising method via sequential decomposition.
\newblock In {\em Proceedings of the IEEE International Symposium on Circuits
  and Systems (ISCAS)}, pages 1--5. IEEE, 2018.

\bibitem{saharia2022image}
Chitwan Saharia, Jonathan Ho, William Chan, Tim Salimans, David~J Fleet, and
  Mohammad Norouzi.
\newblock Image super-resolution via iterative refinement.
\newblock {\em IEEE Transactions on Pattern Analysis and Machine Intelligence},
  45(4):4713--4726, 2022.

\bibitem{singh2014image}
Kuldeep Singh and Rajiv Kapoor.
\newblock Image enhancement using exposure based sub image histogram
  equalization.
\newblock {\em Pattern Recognition Letters}, 36:10--14, 2014.

\bibitem{ueng1995gamma}
Neng-Tsann Ueng and Louis~L Scharf.
\newblock The gamma transform: A local time-frequency analysis method.
\newblock In {\em Conference Record of The Twenty-Ninth Asilomar Conference on
  Signals, Systems and Computers}, volume~2, pages 920--924. IEEE, 1995.

\bibitem{wang2017generative}
Kunfeng Wang, Chao Gou, Yanjie Duan, Yilun Lin, Xinhu Zheng, and Fei-Yue Wang.
\newblock Generative adversarial networks: introduction and outlook.
\newblock {\em IEEE/CAA Journal of Automatica Sinica}, 4(4):588--598, 2017.

\bibitem{wang2013naturalness}
Shuhang Wang, Jin Zheng, Hai-Miao Hu, and Bo Li.
\newblock Naturalness preserved enhancement algorithm for non-uniform
  illumination images.
\newblock {\em IEEE Transactions on Image Processing}, 22(9):3538--3548, 2013.

\bibitem{wang2023ultra}
Tao Wang, Kaihao Zhang, Tianrun Shen, Wenhan Luo, Bjorn Stenger, and Tong Lu.
\newblock Ultra-high-definition low-light image enhancement: A benchmark and
  transformer-based method.
\newblock In {\em Proceedings of the AAAI Conference on Artificial
  Intelligence}, volume~37, pages 2654--2662, 2023.

\bibitem{wei2018deep}
Chen Wei, Wenjing Wang, Wenhan Yang, and Jiaying Liu.
\newblock Deep retinex decomposition for low-light enhancement.
\newblock {\em arXiv preprint arXiv:1808.04560}, 2018.

\bibitem{wu2022uretinex}
Wenhui Wu, Jian Weng, Pingping Zhang, Xu Wang, Wenhan Yang, and Jianmin Jiang.
\newblock Uretinex-net: Retinex-based deep unfolding network for low-light
  image enhancement.
\newblock In {\em Proceedings of the IEEE/CVF Conference on Computer Vision and
  Pattern Recognition}, pages 5901--5910, 2022.

\bibitem{zamir2022restormer}
Syed~Waqas Zamir, Aditya Arora, Salman Khan, Munawar Hayat, Fahad~Shahbaz Khan,
  and Ming-Hsuan Yang.
\newblock Restormer: Efficient transformer for high-resolution image
  restoration.
\newblock In {\em Proceedings of the IEEE/CVF Conference on Computer Vision and
  Pattern Recognition}, pages 5728--5739, 2022.

\bibitem{zhang2018unreasonable}
Richard Zhang, Phillip Isola, Alexei~A Efros, Eli Shechtman, and Oliver Wang.
\newblock The unreasonable effectiveness of deep features as a perceptual
  metric.
\newblock In {\em Proceedings of the IEEE Conference on Computer Vision and
  Pattern Recognition}, pages 586--595, 2018.

\bibitem{zhang2021beyond}
Yonghua Zhang, Xiaojie Guo, Jiayi Ma, Wei Liu, and Jiawan Zhang.
\newblock Beyond brightening low-light images.
\newblock {\em International Journal of Computer Vision}, 129:1013--1037, 2021.

\bibitem{zhang2019kindling}
Yonghua Zhang, Jiawan Zhang, and Xiaojie Guo.
\newblock Kindling the darkness: A practical low-light image enhancer.
\newblock In {\em Proceedings of the ACM International Conference on
  Multimedia}, pages 1632--1640, 2019.

\bibitem{zhu2020zero}
Anqi Zhu, Lin Zhang, Ying Shen, Yong Ma, Shengjie Zhao, and Yicong Zhou.
\newblock Zero-shot restoration of underexposed images via robust retinex
  decomposition.
\newblock In {\em Proceedings of the IEEE International Conference on
  Multimedia and Expo (ICME)}, pages 1--6. IEEE, 2020.

\bibitem{zhu2018generative}
Lin Zhu, Yushi Chen, Pedram Ghamisi, and J{\'o}n~Atli Benediktsson.
\newblock Generative adversarial networks for hyperspectral image
  classification.
\newblock {\em IEEE Transactions on Geoscience and Remote Sensing},
  56(9):5046--5063, 2018.

\end{thebibliography}
}

\end{document}